\documentclass{article}
\usepackage{ijcai17}
\usepackage{graphicx}
\usepackage{amsmath}
\usepackage[small]{caption}
\usepackage{float}
\usepackage{array}
\usepackage{subcaption}
\usepackage{amssymb}
\usepackage{algorithm}
\usepackage{algorithmic}
\usepackage{xcolor}
\usepackage{url}
\usepackage{multirow}
\usepackage{booktabs}

\usepackage{tikz}

\tikzset{
     block/.style={rectangle, draw, fill=none, text width=6em,
                   text centered,minimum height=3em},
     arrow/.style={-{Stealth[]}}
     }
\usetikzlibrary{positioning,arrows.meta}

\DeclareSymbolFont{bbold}{U}{bbold}{m}{n}
\DeclareSymbolFontAlphabet{\mathbbold}{bbold}

\DeclareMathOperator*{\argmin}{arg\,min}

\newcommand{\specialcell}[2][c]{%
  \begin{tabular}[#1]{@{}c@{}}#2\end{tabular}}

\usepackage{times}

\title{COBRA: A Fast and Simple Method for\\ Active Clustering with Pairwise Constraints}
\author{Toon Van Craenendonck, Sebastijan Duman\v{c}i\'{c} \and Hendrik Blockeel\\ 
Department of Computer Science, KU Leuven, Belgium  \\
\{firstname.lastname\}@kuleuven.be}

\begin{document}

\maketitle

\begin{abstract}
Clustering is inherently ill-posed: there often exist multiple valid clusterings of a single dataset, and without any additional information a clustering system has no way of knowing which clustering it should produce. This motivates the use of constraints in clustering, as they allow users to communicate their interests to the clustering system. Active constraint-based clustering algorithms select the most useful constraints to query, aiming to produce a good clustering using as few constraints as possible. We propose COBRA, an active method that first over-clusters the data by running K-means with a $K$ that is intended to be too large, and subsequently merges the resulting small clusters into larger ones based on pairwise constraints. In its merging step, COBRA is able to keep the number of pairwise queries low by maximally exploiting constraint transitivity and entailment. We experimentally show that COBRA outperforms the state of the art in terms of clustering quality and runtime, without requiring the number of clusters in advance.  
\end{abstract}

\section{Introduction}
Clustering is inherently subjective \cite{Caruana06metaclustering,ScienceOrArt}: a single dataset can often be clustered in multiple ways, and different users may prefer different clusterings. This subjectivity is one of the motivations for constraint-based (or semi-supervised) clustering \cite{Wagstaff01constrainedK-means,Bilenko2004}. Methods in this setting exploit background knowledge to obtain clusterings that are more aligned with the user's preferences. Often, this knowledge is given in the form of pairwise constraints that indicate whether two instances should be in the same cluster (a must-link constraint) or not (a cannot-link constraint) \cite{Wagstaff01constrainedK-means}. In traditional constraint-based clustering systems the set of constraints is assumed to be given a priori, and in practice, the pairs that are queried are often selected randomly. In contrast, in active clustering \cite{basu:sdm04,Mallapragada2008,Xiong2014} it is the method itself that decides which pairs to query. Typically, active methods query pairs that are more informative than random ones, which improves clustering quality.

This work introduces an active constraint-based clustering method named \textit{Constraint-Based Repeated Aggregation} (COBRA).
It differs from existing approaches in several ways.
First, it aims to maximally exploit constraint transitivity and entailment \cite{Wagstaff01constrainedK-means}, two properties that allow deriving additional constraints from a given set of constraints. 
By doing this, the actual number of pairwise constraints that COBRA works with is typically much larger than the number of pairwise constraints that are queried from the user.
Secondly, COBRA introduces the assumption that there exist small local regions in the data that are grouped together in all potential clusterings. To clarify this, consider the example of clustering images of people taking different poses (e.g. facing left or right). There are at least two natural clustering targets for this data: one might want to cluster based on identity or pose. In an appropriate feature space, one expects images that agree on both criteria (i.e.\ of a single person, taking a single pose) to be close. There is no need to consider all of these instances individually, as they will end up in the same cluster for each of the two targets that the user might be interested in. COBRA aims to group such instances into a \textit{super-instance}, such that they can be treated jointly in the clustering process. Doing so substantially reduces the number of pairwise queries.
Thirdly, COBRA is an inherently active method: the constraints are selected during the execution of the algorithm itself, as constraint selection and algorithm execution are intertwined. 
In contrast, existing approaches consist of a component that selects constraints and another one that uses them during clustering. 

Our experiments show that COBRA outperforms state-of-the-art active clustering methods in terms of both clustering quality and runtime. Furthermore, it has the distinct advantage that it does not require knowing the number of clusters beforehand, as the competitors do. In many realistic clustering scenarios this number is not known, and running an algorithm with the wrong number of clusters often results in a significant decrease in clustering quality. 

We discuss related work on (active) constraint-based clustering in section \ref{sec:related}. In section \ref{sec:cobra} we elaborate the key ideas in COBRA and describe the method in more detail. We present our experimental evaluation in section \ref{sec:results}, and conclude in section \ref{sec:conclusion}.

\section{Background and Related Work}
\label{sec:related}
Most existing constraint-based methods are extensions of well-known unsupervised clustering algorithms. They use the constraints either in an adapted clustering procedure \cite{Wagstaff01constrainedK-means,Rangapuram2012,Wang2014}, to learn a similarity metric \cite{xing2002distance,Davis:2007:IML:1273496.1273523}, or both \cite{Bilenko2004,probframework}. Constraint-based extensions have been developed for most clustering algorithms, including K-means \cite{Wagstaff01constrainedK-means,Bilenko2004}, spectral clustering \cite{Rangapuram2012,Wang2014}, DBSCAN \cite{SSDBSCAN,Campello} and EM \cite{Shental2004}. 

Basu et al.\ \shortcite{basu:sdm04} introduce a strategy to select the most informative constraints, prior to performing a single run of a constraint-based clustering algorithm. They show that active constraint selection can improve clustering performance. Several selection strategies have been proposed since \cite{Mallapragada2008,Xu2005,Xiong2014}, most of which are based on the classic approach of uncertainty sampling. As COBRA also chooses which pairs to query, we consider it to be an active method, and in our experiments we compare to other methods in this setting. Note, however, that COBRA is quite different from existing methods in active constrained clustering and active learning in general. For COBRA, selecting which pairs to query is inherent to the clustering procedure, whereas for most other methods the selection strategy is optional and considered to be a separate component.

In its core, COBRA is related to hierarchical clustering as it follows the same procedure of sequentially trying to merge the two closest clusters.
Constraints have been used in hierarchical clustering before but in different ways. 
Davidson et al.\ \shortcite{Davidson2009}, for example, present an algorithm to find a clustering hierarchy that is consistent with a \emph{given} set of constraints. 
Nogueira et al.\ \shortcite{Nogueira2012} propose an active semi-supervised hierarchical clustering algorithm that is based on merge confidence.  
Also related to ours is the work of Campello et al.\ \shortcite{Campello}, who have developed a framework to extract from a given hierarchy a flat clustering that is consistent with a given set of constraints.
The key difference is that COBRA starts from super-instances, i.e.\ small clusters produced by K-means, and that each merging decision is settled by a pairwise constraint. 

The idea of working with a small number of representatives (in our case the super-instance medoids, as will be discussed in section \ref{sec:cobra}) instead of all individual instances has been used before, but for very different purposes. For example, Yan et al.\ \shortcite{Yan:2009:FAS:1557019.1557118} use it to speed up unsupervised spectral clustering, whereas we use it to reduce the number of pairwise queries.

\section{Constraint-Based Repeated Aggregation}
\label{sec:cobra}


\begin{figure*}[ht]
\centering
\begin{subfigure}{.23\textwidth}
  \centering
  \includegraphics[width=0.85\linewidth]{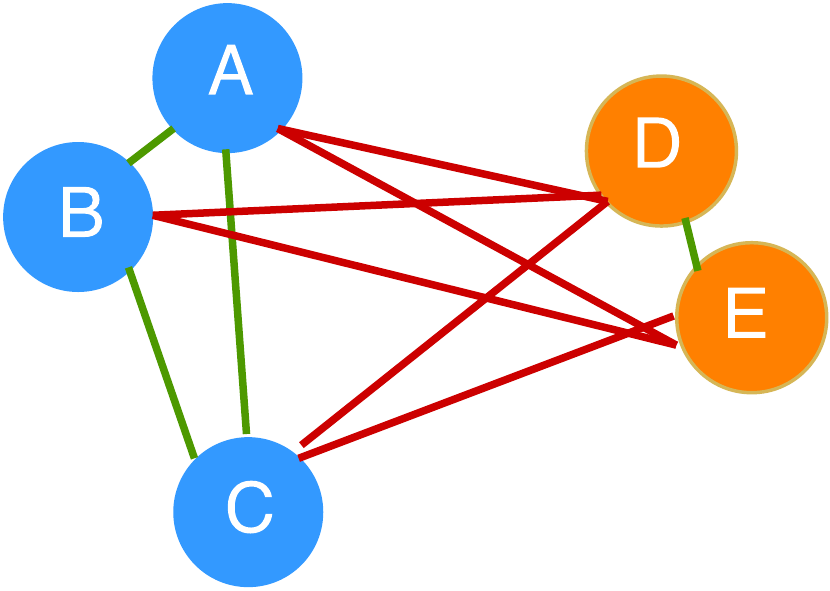}
  \caption{}
  \label{fig:sub1}
\end{subfigure}%
\hspace{0.4em}
\begin{subfigure}{.23\textwidth}
  \centering
  \includegraphics[width=0.85\linewidth]{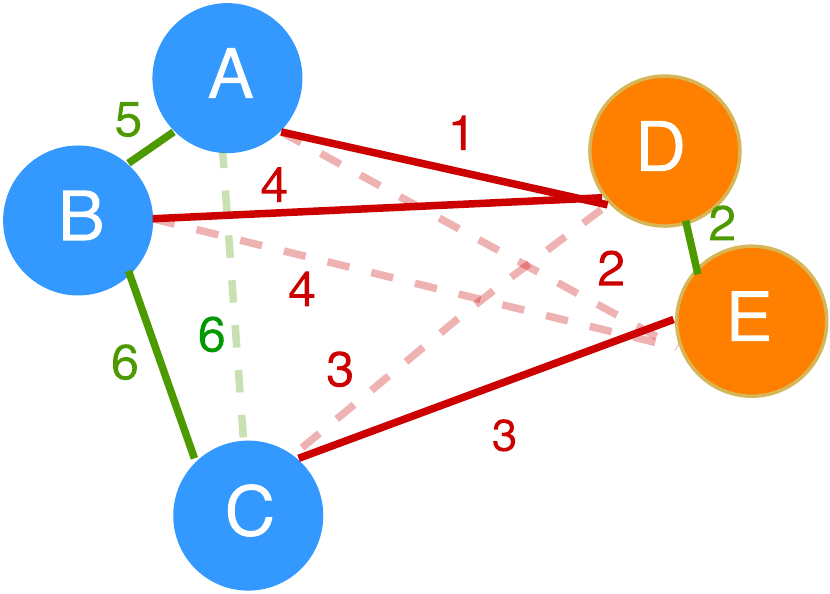}
  \caption{}
  \label{fig:sub2}
\end{subfigure}%
\hspace{0.4em}
\begin{subfigure}{.23\textwidth}
  \centering
  \includegraphics[width=0.85\linewidth]{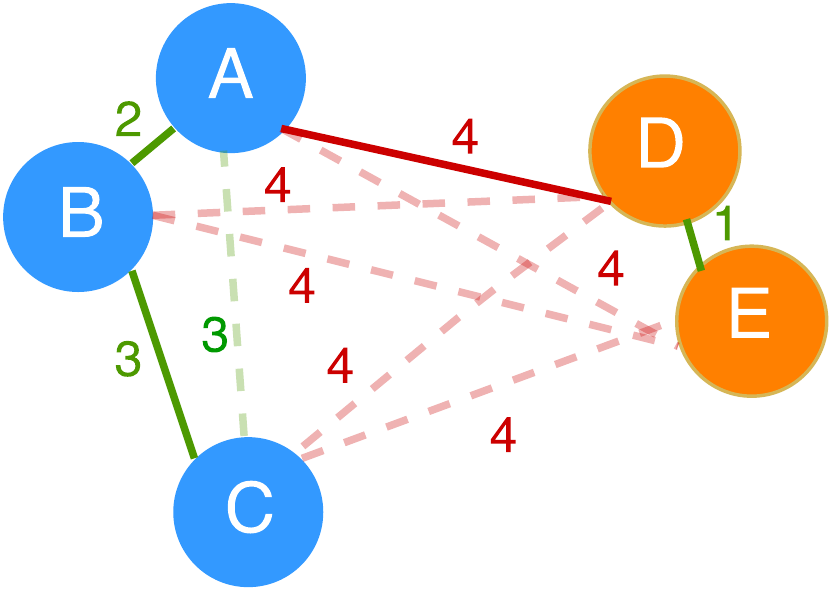}
  \caption{}
  \label{fig:sub3}
\end{subfigure}%
\hspace{0.4em}
\begin{subfigure}{.23\textwidth}
  \centering
  \includegraphics[width=0.85\linewidth]{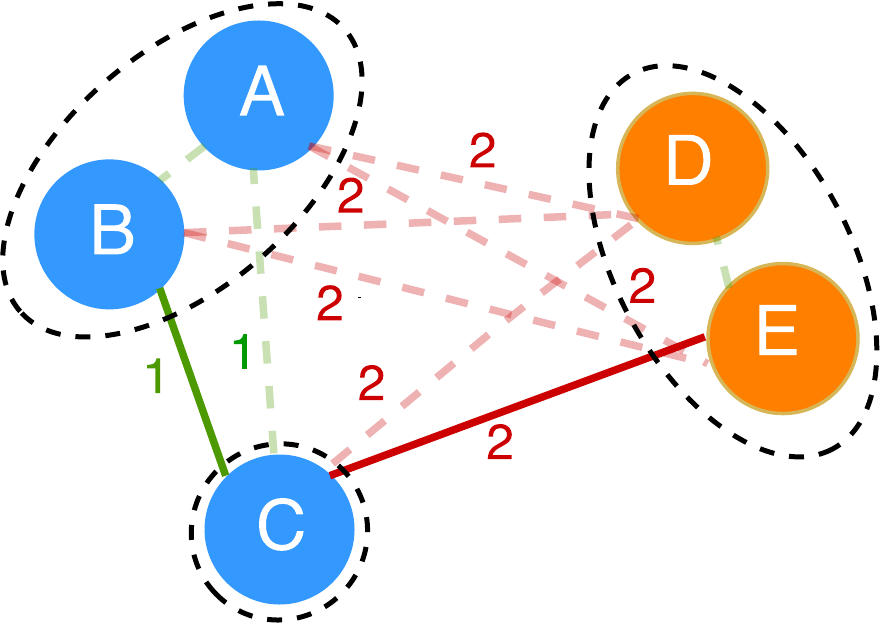}
  \caption{}
  \label{fig:sub4}
\end{subfigure}

\caption{Illustration of different querying strategies. Colors indicate the desired clustering. Solid green lines indicate \textit{must-link}, red ones \textit{cannot-link} constraints. Dashed lines indicate derived constraints with the same color code. The number next to the solid line indicates the ordering of queried constraints, whereas the number next to the dashed line indicates the constraint number from which the constraint was derived. (a) Querying all 10 constraints (b) Exploiting entailment and transitivity results in querying 6 constraints. (c) Querying the closest pairs first results in 4 constraints.  (d) Introducing \textit{super-instances} (dashed ellipses) results in only 2 queries. }
\label{fig:cobra_example}
\end{figure*}

Constraint-based clustering algorithms aim to produce a clustering of a dataset that resembles an unknown target clustering $Y$ as close as possible. The algorithm cannot query the cluster labels in $Y$ directly, but can query the relation between pairs of instances. A must-link constraint is obtained if the instances have the same cluster label in $Y$, a cannot-link constraint otherwise. The aim is to produce a clustering that is close to the target clustering $Y$, using as few pairwise queries as possible. 

Several strategies can be used to exploit constraints in clustering. 
Figure~\ref{fig:cobra_example} illustrates some of them.
The most naive strategy is to query all pairwise relations, and construct clusters as sets of instances that are connected by a must-link constraint (Figure~\ref{fig:sub1}).
Though this is clearly not a good strategy in any scenario, it allows us to formulate a baseline for further improvements.
It always results in a perfect clustering, but at a very high cost: for a dataset of $N$ instances, $\binom N 2$ questions are asked.

The previous strategy can be improved by exploiting \textit{constraint transitivity} and \textit{entailment}, two well known properties in constraint-based clustering \cite{Wagstaff01constrainedK-means,Bilenko2004}.
Must-link constraints are known to be transitive: 
\begin{center}
$\texttt{must-link}(A,B) \wedge \texttt{must-link}(B,C) \Rightarrow \texttt{must-link}(A,C),$
\end{center}
whereas cannot-link constraints have an entailment property:
\begin{center}
 $\texttt{must-link}(A,B) \wedge \texttt{cannot-link}(B,C) \Rightarrow \texttt{cannot-link}(A,C).$
\end{center}
Thus, every time a constraint is queried and added to the set of constraints, transitivity and entailment can be applied to expand the set.
This strategy is illustrated in Figure~\ref{fig:sub2}. 
Exploiting transitivity and entailment significantly reduces the number of pairwise queries needed to obtain a clustering, without a loss in clustering quality.

The order in which constraints are queried strongly influences the number of constraints that can be derived.
In general, it is better to obtain must-link constraints early on.
That way, any future query involving one of the instances connected by a must-link also applies to all others. This suggests querying the closest pairs first, as they are more likely to belong to the same cluster and hence be connected by a must-link constraint. This is strategy is illustrated in Figure~\ref{fig:sub3}.

The previous strategies all obtain a perfect clustering, but require a high number of queries which makes them inapplicable for reasonably sized datasets. 
To further reduce the number of queries, COBRA groups similar instances into \textit{super-instances} and only clusters their representatives, i.e. medoids.
It assumes that all instances within a super-instance are connected by a must-link constraint.
While clustering the medoids, COBRA uses both previously discussed strategies of querying the closest pairs and exploiting transitivity and entailment.
This strategy, illustrated in Figure~\ref{fig:sub4}, results in a substantial reduction of the number of queries.
It does not always result in a perfect clustering as it is possible that the instances within a particular super-instance should not be grouped together w.r.t. the target clustering.

Table~\ref{table:questions_counts} illustrates to what extent each of the improvements described above reduces the number of queries. 
We perform an extensive evaluation of the quality of the clusterings that COBRA produces in section \ref{sec:results}.

\begin{table*}[ht]
\centering
\begin{tabular}{ccc}
 	\toprule
   \textbf{dataset} & \textbf{\# instances} &  \textbf{\specialcell{total \# pairs}}   \\ \midrule
iris &  147 & 10731     \\
wine &  178 & 15753   \\
dermatology & 358 &  63903     \\
hepatitis & 112 & 6216    \\
ecoli & 336 & 56280    \\
opdigits389 & 1151 & 661825 \\
	\bottomrule
	\end{tabular}	
$\xrightarrow[\text{entailment}]{\text{transitivity}}$
    \begin{tabular}{c}
 \toprule
 \textbf{\specialcell{\# queries}}   \\ \midrule 
 409     \\
 457   \\
  1660     \\
 173    \\
1343    \\
3380 \\
	\bottomrule
	\end{tabular}	
    $\xrightarrow[\text{first}]{\text{closest pairs}}$
    \begin{tabular}{c}
 \toprule
 \textbf{\specialcell{\# queries}}   \\ \midrule 
 155     \\
 187   \\
  379     \\
 140    \\
440    \\
1164 \\
	\bottomrule
	\end{tabular}	
        $\xrightarrow[\text{instances}]{\text{super}}$
    \begin{tabular}{c}
 \toprule
 \textbf{\specialcell{\# queries}}   \\ \midrule 
 34     \\
 35   \\
  42     \\
 39    \\
51    \\
54 \\
	\bottomrule
	\end{tabular}	
 \caption{This table shows the total number of pairwise relations in several datasets (column 3), as well the number of pairwise queries that is required when exploiting transitivity and entailment (column 4, these numbers are averages of runs for 5 random orderings of pairwise queries), and additionally querying the closest pairs first (column 5). The last column shows the number of pairwise queries (averaged over 5 runs) when COBRA is run with 25 super-instances.} \label{table:questions_counts}
\end{table*}

\subsection{Algorithmic Description}
\label{sec:cobra_algo}


\begin{algorithm}[ht]
\caption{COBRA}
\label{algo:COBRA}
\begin{algorithmic}[1]
 \REQUIRE $\mathcal{X}$: a dataset, \\ 
         \quad \ \ \ \ \ \  $N_{S}$: the number of super-instances, \\ 
\ENSURE $\mathbf{C}$, a clustering of $D$

\STATE $\{\mathcal{S}_i\}_{i=1}^{N_S} = \texttt{K-means}(\mathcal{X},N_{S})$
\STATE $\forall i = 1, \ldots, N_S: \mathcal{C}_i = \{\mathbf{s}_i\} $
\STATE $\mathcal{CL} = \emptyset $

\STATE $mergeHappened = True$
\WHILE  {$mergeHappened$}
	\STATE $\mathcal{P} = \{ \mathcal{C}_1,\mathcal{C}_2 : \nexists \mathbf{x} \in \mathcal{C}_1,\mathbf{y} \in \mathcal{C}_2: (\mathbf{x},\mathbf{y}) \in \mathcal{CL}   \}$, ordered by $d(\mathcal{C}_1,\mathcal{C}_2)$
	\STATE $mergeHappened = False$
	\FOR {$\mathcal{C}_1 , \mathcal{C}_2$ in $\mathcal{P}$}
		\STATE $\mathbf{s}_a , \mathbf{s}_b = \argmin_{\mathbf{s}_1 \in \mathcal{C}_1, \mathbf{s}_2 \in \mathcal{C}_2} \lVert \mathbf{s}_1 - \mathbf{s}_2 \rVert_{2}$
        \IF {$\texttt{must-link}(\mathbf{s}_a , \mathbf{s}_b)$}
			\STATE $\mathcal{C}_1  = \mathcal{C}_1 \cup \mathcal{C}_2$
			\STATE $\mathbf{C} = \mathbf{C} \setminus \{\mathcal{C}_2\}$
			\STATE $mergeHappened = True$
			\STATE break
        	\ELSE
        		\STATE $\mathcal{CL} = \mathcal{CL} \cup \{ (\mathbf{s}_a , \mathbf{s}_b\})$
        	\ENDIF
    \ENDFOR 
\ENDWHILE
\RETURN $\mathbf{C}$
\end{algorithmic}    
\end{algorithm}

After presenting the main motivations for each step of COBRA, we now give a more detailed description in Algorithm \ref{algo:COBRA}. 
Let $\mathcal{X} = \{x_i\}_{i=1}^N, x_i \in \mathbb{R}^{m}$ the instances to be clustered.
The set of instances $\mathcal{X}$ is first \textit{over-clustered} into $N_S$ disjoint subsets, namely super-instances $\{\mathcal{S}_i\}^{N_S}_{i=1}$ , such that $\bigcup _{i} \mathcal{S}_i = \mathcal{X}$. This over-clustering is obtained by running K-means with a $K$ that may be significantly larger than the actual number of clusters.
Each super-instance $\mathcal{S}_i$ is represented by its medoid $\mathbf{s}_i$, forming a set of super-instance representatives $\mathbf{S} = \{\mathbf{s}_1,\ldots,\mathbf{s}_{N_S}\}$. 
All pairwise queries that are performed are between these super-instance representatives.
The goal of COBRA is now to cluster these representatives into disjoint subsets $\{\mathcal{C}_1,\ldots,\mathcal{C}_{N_C}\}$ of $\mathbf{S}$ ($\bigcup _{i} \mathcal{C}_i = \mathbf{S}$).

Each cluster $\mathcal{C}_i$ is a set of super-instance representatives, but conceptually contains all the points in the corresponding super-instances. 
The number of clusters $N_C$ is unknown a priori and will be determined during the clustering procedure.  
Initially, there are $N_S$ clusters, each containing a single super-instance representative, as shown on line 2 in Algorithm \ref{algo:COBRA}. These clusters are merged (if necessary) in the subsequent while loop.

In each iteration, COBRA first sorts all pairs of clusters between which there is no cannot-link constraint (line 6).  The distance between clusters, by which the pairs are sorted, is defined as follows (as in single-linkage clustering): 

\begin{equation}
	d(\mathcal{C}_1,\mathcal{C}_2) = \min_{\mathbf{s}_1 \in \mathcal{C}_1,\mathbf{s}_2 \in \mathcal{C}_2} \lVert \mathbf{s}_1 - \mathbf{s}_2 \rVert_{2} 
\end{equation}

Next, COBRA loops over the pairs of clusters and checks whether they should be merged. 
It starts by selecting the closest pair of super-instance representatives of the clusters (line 9), and asks whether they should be in the same cluster (line 10). 
If this is the case, the clusters are merged (lines 11 and 12), and the while-loop is restarted (as the set of clusters is changed). 
If that is not the case, the pair of representatives is added to the set of cannot-link constraints (line 16), and the inner loop continues by inspecting the next pair of clusters. 
The execution stops when all clusters are complete and no merge is to be done anymore.

\subsubsection*{Number of Super-instances and Number of Queries }

\begin{figure}[ht]
\centering
  \centering
  \includegraphics[width=0.8\linewidth]{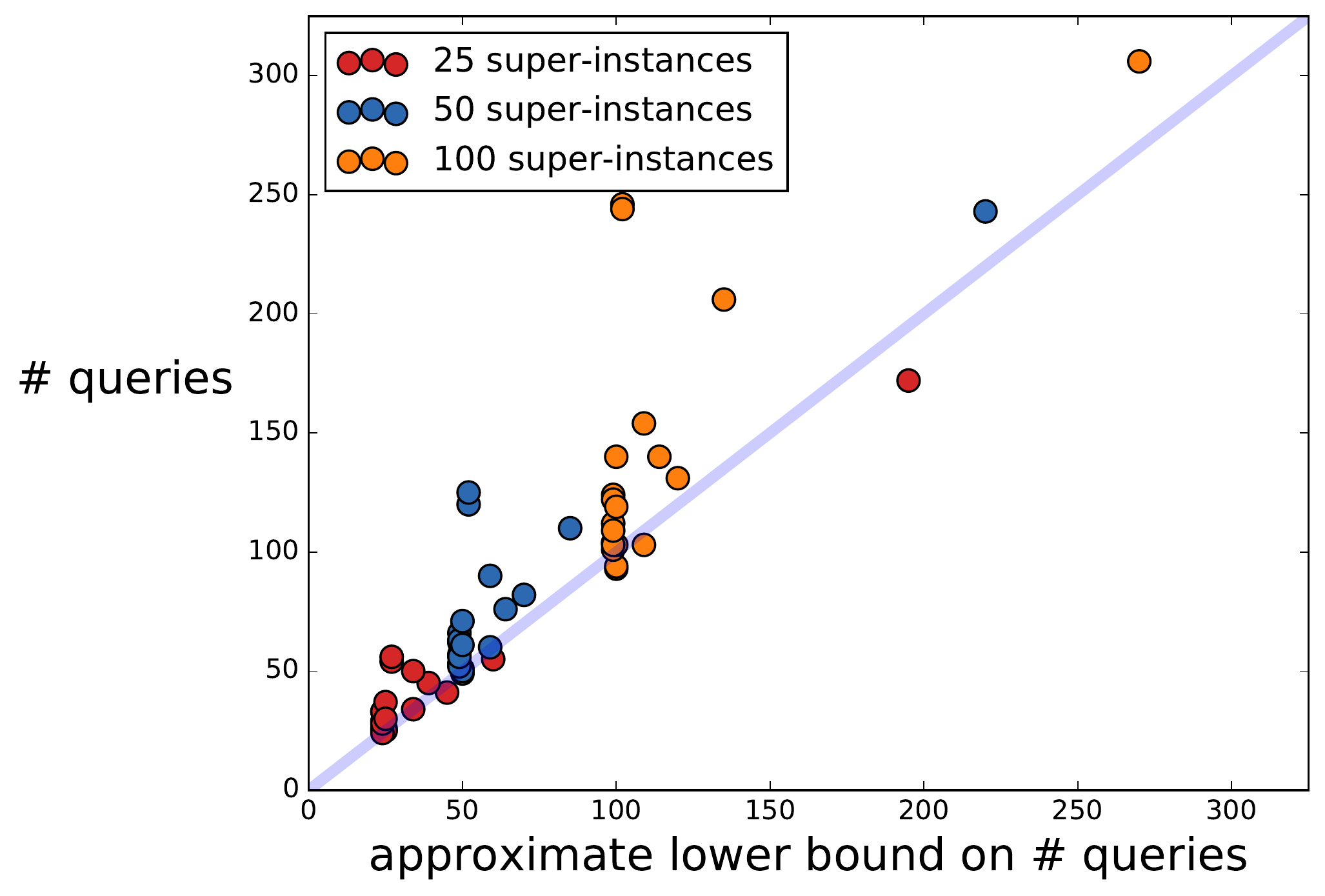}
  \caption{The estimated lower bound on the number of queries vs. the number of queries needed by COBRA (averaged over 5 CV folds). Each dot corresponds to one of the 21 clustering tasks.}
  \label{fig:estimate_vs_real}
\end{figure}

The exact number of queries COBRA needs depends on the extent to which querying the closest pairs first leads to must-link constraints and on the actual number of clusters.
It is thus difficult to determine it before the execution.
However, an estimate in terms of a lower and upper bound can be posed:

\begin{center}
	$N_S - N_C + \binom{N_C}{2} \lesssim $ \# queries $\leq \binom{N_S}{2}$
\end{center}
with $N_S$ the number of super-instances, and $N_C$ the number of clusters in the target clustering.
In the worst case, COBRA will need to query all pairwise relations between super-instances, which requires $\binom{N_S}{2}$ queries. This happens if there is a cannot-link between each pair of super-instances.
In practice, COBRA typically needs much less than $\binom{N_S}{2}$ queries.
If (i) super-instances are perfectly homogeneous w.r.t.\ the target clustering and (ii) the distances between must-link pairs are smaller than the distances between cannot-link pairs, COBRA needs exactly $N_S - N_C + \binom{N_C}{2}$ queries.
$N_S-N_C$ must-link constraints are needed to merge the $N_S$ super-instances into the $N_C$ clusters, followed by an additional $\binom{N_C}{2}$ queries to ensure nothing can be merged anymore.
This formula is inapplicable in practice as the number of clusters is not known beforehand, and thus serves as a means to understand the number of queries that might be needed.
Figure \ref{fig:estimate_vs_real} compares this lower bound to the actual number of queries that was needed by COBRA for 21 clustering tasks (these will be described in more detail in section \ref{sec:results}).
For most tasks, the actual number of pairwise queries is relatively close to the lower bound. 
 It is possible to get a smaller number of queries than that suggested by the lower bound: this happens if a single super-instance contains the instances of two actual clusters, rendering the $\binom{N_C}{2}$ factor inaccurate.

\begin{table*}\footnotesize
\centering
  \caption{Wins and losses aggregated over all 21 clustering tasks. After each win (loss) count, we report the average margin by which COBRA wins (loses). For win counts marked with an asterisk, the differences are significant according to the Wilcoxon test with $p<0.05$.    } \label{table:winloss}
  \begin{tabular}{c  c  c  c  c  c  c  }
  \toprule
\multicolumn{1}{c}{} & \multicolumn{2}{c}{25 super-instances} & \multicolumn{2}{c}{50 super-instances}  & \multicolumn{2}{c}{100 super-instances}  \\ 
\multicolumn{1}{c}{}  & win & loss & win & loss & win & loss \\ 
\midrule
COBRA vs. MPCKM-MinMax & \textbf{13} (0.14) & 8 (0.12) &  \textbf{13} (0.16) & 8 (0.09) &  \textbf{12} (0.19) & 9 (0.05) \\
COBRA vs. MPCKM-NPU & \textbf{11} (0.16) & 10 (0.11) &  \textbf{17*} (0.12) & 4 (0.09) &  \textbf{12} (0.17) & 9 (0.06) \\
COBRA vs. COSC-MinMax & \textbf{15*} (0.21) & 6 (0.05) &  \textbf{16*} (0.21) & 5 (0.06) &  \textbf{14*} (0.21) & 7 (0.04)  \\
COBRA vs. COSC-NPU& \textbf{15*} (0.20) & 6 (0.04) &  \textbf{14*} (0.23) & 7 (0.04) &  \textbf{13*} (0.23) & 8 (0.03) \\
 \bottomrule
  \end{tabular}
\end{table*}

\section{Experimental Evaluation}
\label{sec:results}

In this section, we discuss the experimental evaluation of COBRA.

\subsubsection*{Existing Constraint-based Algorithms}
We compare COBRA to the following state-of-the-art constraint-based clustering algorithms:

\begin{itemize}
	\item MPCKMeans \cite{Bilenko2004} is a hybrid constraint-based extension of K-means: it uses metric learning and an adapted clustering procedure combining the within-cluster sum of squares with the cost of violating constraints in the objective. We use the implementation that is available in the WekaUT package\footnote{{\scriptsize \url{http://www.cs.utexas.edu/users/ml/risc/code/}}}.
	\item Constrained Spectral Clustering (COSC) \cite{Rangapuram2012} is based on spectral clustering, but optimizes for a modified objective that also takes constraint violation into account. We use the code provided by the authors on their web page\footnote{{\scriptsize \url{http://www.ml.uni-saarland.de/code/cosc/cosc.htm}}}. 
\end{itemize}

It is important to note that, in contrast to COBRA, COSC and MPCKMeans require the number of clusters as an input parameter. In our experiments, the true number of clusters is provided to these algorithms. In many clustering applications, however, this number is typically not known beforehand. Thus, COSC and MPCKMeans are at an advantage.

\subsubsection*{Active Selection Strategies}

Each of these algorithms is combined with the following two active selection strategies:

\begin{itemize}
	\item MinMax \cite{Mallapragada2008} starts with an \emph{exploration} phase in which $K$ (the number of clusters, assumed to be known in advance) neighborhoods with cannot-links between them are sought. In the subsequent \emph{consolidation} phase these neighborhoods are expanded by selecting the most uncertain instances and determining their neighborhood membership by means of pairwise constraints. We set the width parameter of the Gaussian kernel to the 20th percentile of the distribution of pairwise Euclidean distances, as in \cite{Mallapragada2008}.
	\item NPU \cite{Xiong2014} is also based on the concept of neighborhoods, but in contrast to MinMax it is an iterative method: the data is clustered multiple times, and each clustering is used to determine the next set of pairs to query. Xiong et al.\ \shortcite{Xiong2014} show that NPU typically outperforms MinMax in terms of clustering quality, at the cost of increased runtime.   
\end{itemize}

\subsubsection*{Datasets}

We experiment with 15 UCI datasets: iris, wine, dermatology, hepatitis, glass, ionosphere, optdigits389, ecoli, breast-cancer-wisconsin, segmentation, column\_2C, parkinsons, spambase, sonar and yeast. Most of these datasets have been used in earlier work on constraint-based clustering \cite{Bilenko2004,Xiong2014}. Optdigits389 contains digits 3, 8 and 9 of the UCI handwritten digits data \cite{Bilenko2004,Mallapragada2008}. Duplicate instances were removed for all these datasets, and all data was normalized between 0 and 1. We also perform experiments on the CMU faces dataset, which contains 624 images of 20 persons taking different poses, with different expressions, with and without sunglasses. Hence, this dataset has 4 target clusterings: identity, pose, expression and sunglasses. We extract a 2048-value feature vector for each image by running it through the pre-trained Inception-V3 network \cite{inceptionnet} and storing the output of the second last layer. Finally, we also cluster the 20 newsgroups text data. For this dataset, we consider two tasks: clustering documents from 3 newsgroups on related topics (the target clusters are comp.graphics, comp.os.ms-windows and comp.windows.x, as in \cite{basu:sdm04,Mallapragada2008}), and clustering documents from 3 newsgroups on very different topics (alt.atheism, rec.sport.baseball and sci.space, as in \cite{basu:sdm04,Mallapragada2008}). We first extract tf-idf features, and next apply latent semantic indexing (as in \cite{Mallapragada2008}) to reduce the dimensionality to 10. This brings the total to 17 datasets, for which 21 clustering tasks are defined (15 UCI datasets with a single target, CMU faces with 4 targets, and 2 subsets of the 20 newsgroups data).

\subsubsection*{Experimental Methodology}
We use a cross-validation procedure that is highly similar to the ones used in e.g.\ \cite{basu:sdm04} and \cite{Mallapragada2008}. In each of 5 folds, 20\% of the instances are set aside as the test set. The clustering algorithm is then run on the entire dataset, but can only query pairwise constraints for which both instances are in the training set. To evaluate the quality of the resulting clustering, we compute the Adjusted Rand index (ARI, \cite{ARI}) only on the instances in the test set. The ARI measures the similarity between two clusterings, in this case between the one produced by the constraint-based clustering algorithm and the one indicated by the class labels. An ARI of 0 means that the clustering is not better than random, 1 indicates a perfect clustering. The final score for an algorithm for a particular dataset is computed as the average ARI over the 5 folds. 

The exact number of pairwise queries is not known beforehand for COBRA, but more super-instances generally results in more queries. To evaluate COBRA with varying amounts of user input, we run it with 25, 50 and 100 super-instances.  For each fold, we execute the following steps:
\begin{itemize}
	\item Run COBRA and count how many constraints it needs.
    \item Run the competitors with the same number of constraints.
    \item Evaluate the resulting clusterings by computing the ARI on the test set.
\end{itemize}
To make sure that COBRA only queries pairs of which both instances are in the training set, the medoid of a super-instance is calculated based on only the training instances in that super-instance (and as such, test instances are never queried during clustering). In the rare event that a super-instance contains only test instances, it is merged with the nearest super-instance that does contain training instances. For the MinMax and NPU selection strategies, pairs involving an instance from the test set are simply excluded from selection.

\subsection*{Results}
 
The results over all 21 clustering tasks are summarized in Tables \ref{table:winloss} and \ref{table:ranks}. Table \ref{table:winloss} reports wins and losses against each of the 4 competitors. It shows that COBRA tends to produce better clusterings than its competitors. The difference with COSC is significant according to the Wilcoxon test with $p<0.05$, whereas the difference with MPCKMeans is not. Table \ref{table:ranks} shows the average ranks for COBRA and its competitors. The Friedman aligned rank test \cite{AlignedRank}, which has more power than the Friedman test when the number of algorithms under comparison is low \cite{StatTests}, indicates that for 50 and 100 super-instances, the differences in rank between COBRA and all competitors are significant, using a posthoc Holm test with $p<0.05$.

\begin{table}[h!]\scriptsize
  \caption{For each dataset, all algorithms are ranked from 1 (best) to 5 (worst). This table shows the average ranks for 25, 50 and 100 super-instances. Algorithms for which the difference with COBRA is significant according to the Friedman aligned rank test and a post-hoc Holm test with $p<0.05$ are marked with an asterisk. }\label{table:ranks}
  \begin{tabular}{ c  c  c  c  c  c }
  \toprule
\multicolumn{2}{c}{25 super-instances} & \multicolumn{2}{c}{50 super-instances} & \multicolumn{2}{c}{100 super-instances} \\ 
\midrule
COBRA & 2.43 & COBRA & 2.14 & COBRA & 2.52 \\
MPCK-NPU & 3.00 & MPCK-MM* & 3.00 & COSC-NPU* & 2.98 \\
MPCK-MM & 3.07 & COSC-NPU* & 3.02 & MPCK-NPU* & 3.00\\
COSC-MM* & 3.12 & COSC-MM* & 3.26 & MPCK-MM* & 3.19 \\
COSC-NPU* & 3.40 & MPCK-NPU* & 3.57 & COSC-MM* & 3.31 \\
 \bottomrule
  \end{tabular} 
  \end{table}

\begin{figure*}[ht]
\centering
\begin{subfigure}{.23\textwidth}
  \centering
  \includegraphics[width=1.0\linewidth]{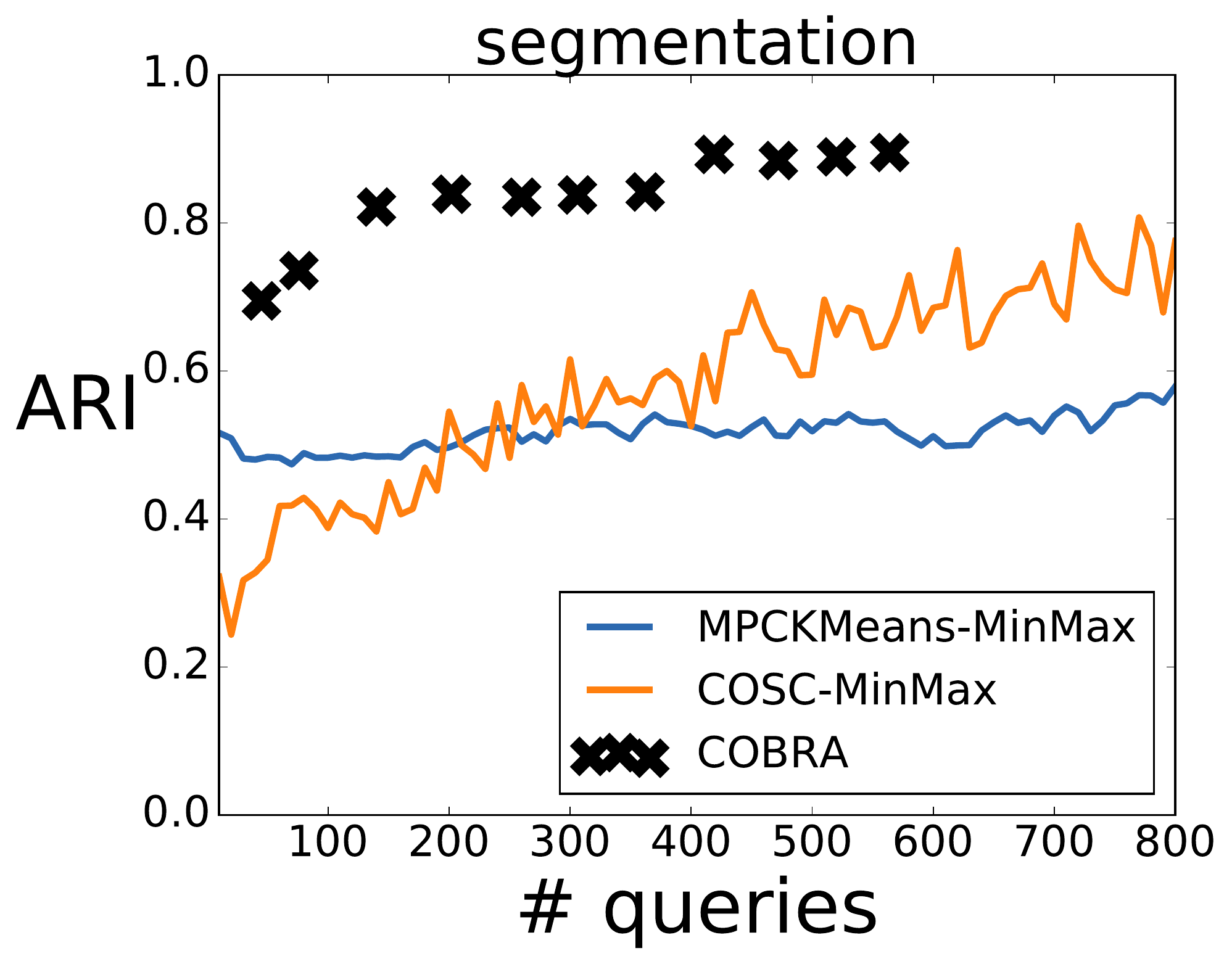}
  \caption{}
  \label{fig:n_constraints_1}
\end{subfigure}%
\begin{subfigure}{.23\textwidth}
  \centering
  \includegraphics[width=1.0\linewidth]{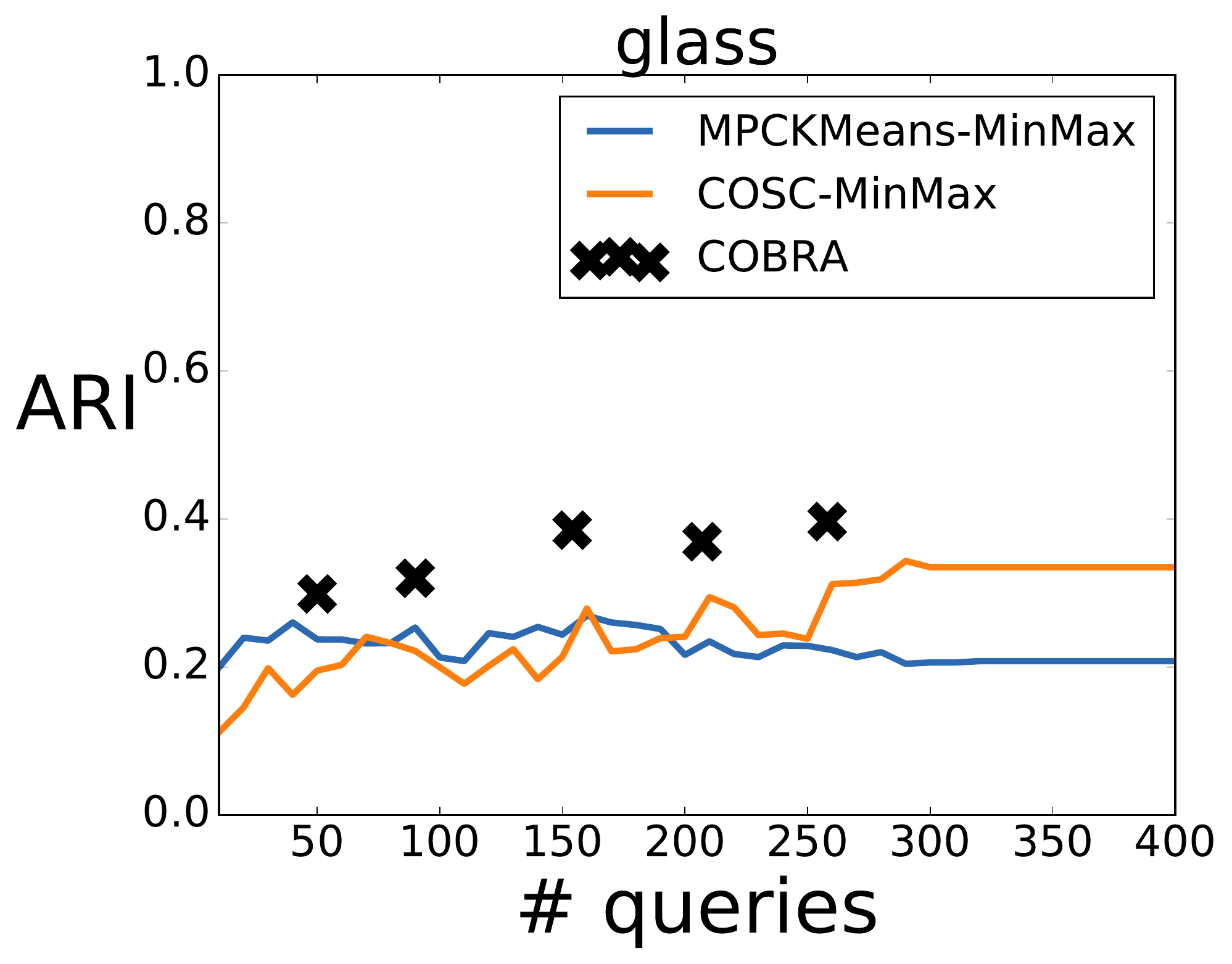}
  \caption{}
  \label{fig:n_constraints_2}
\end{subfigure}%
\begin{subfigure}{.23\textwidth}
  \centering
  \includegraphics[width=1.0\linewidth]{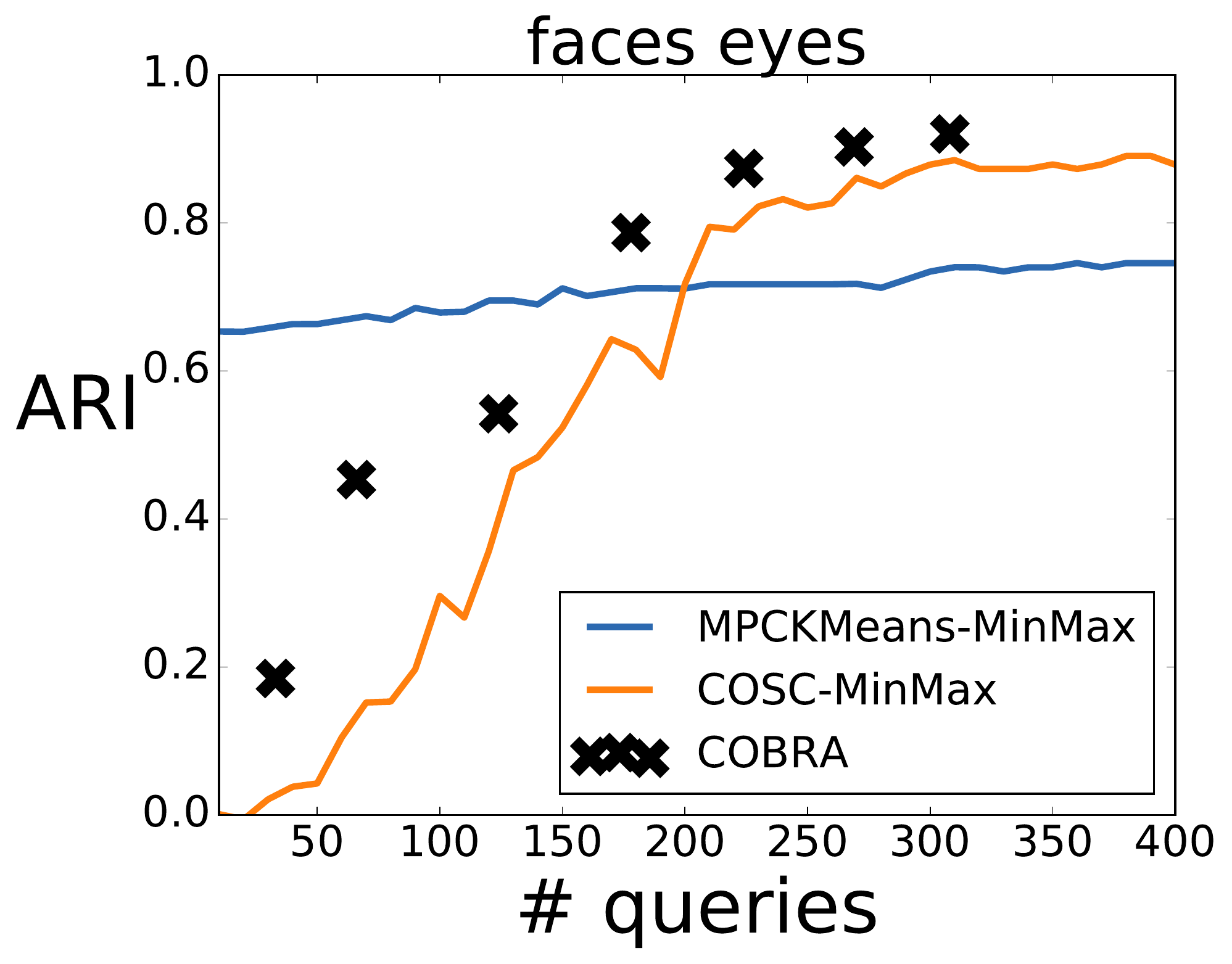}
  \caption{}
  \label{fig:n_constraints_3}
\end{subfigure}%
\begin{subfigure}{.23\textwidth}
  \centering
  \includegraphics[width=1.0\linewidth]{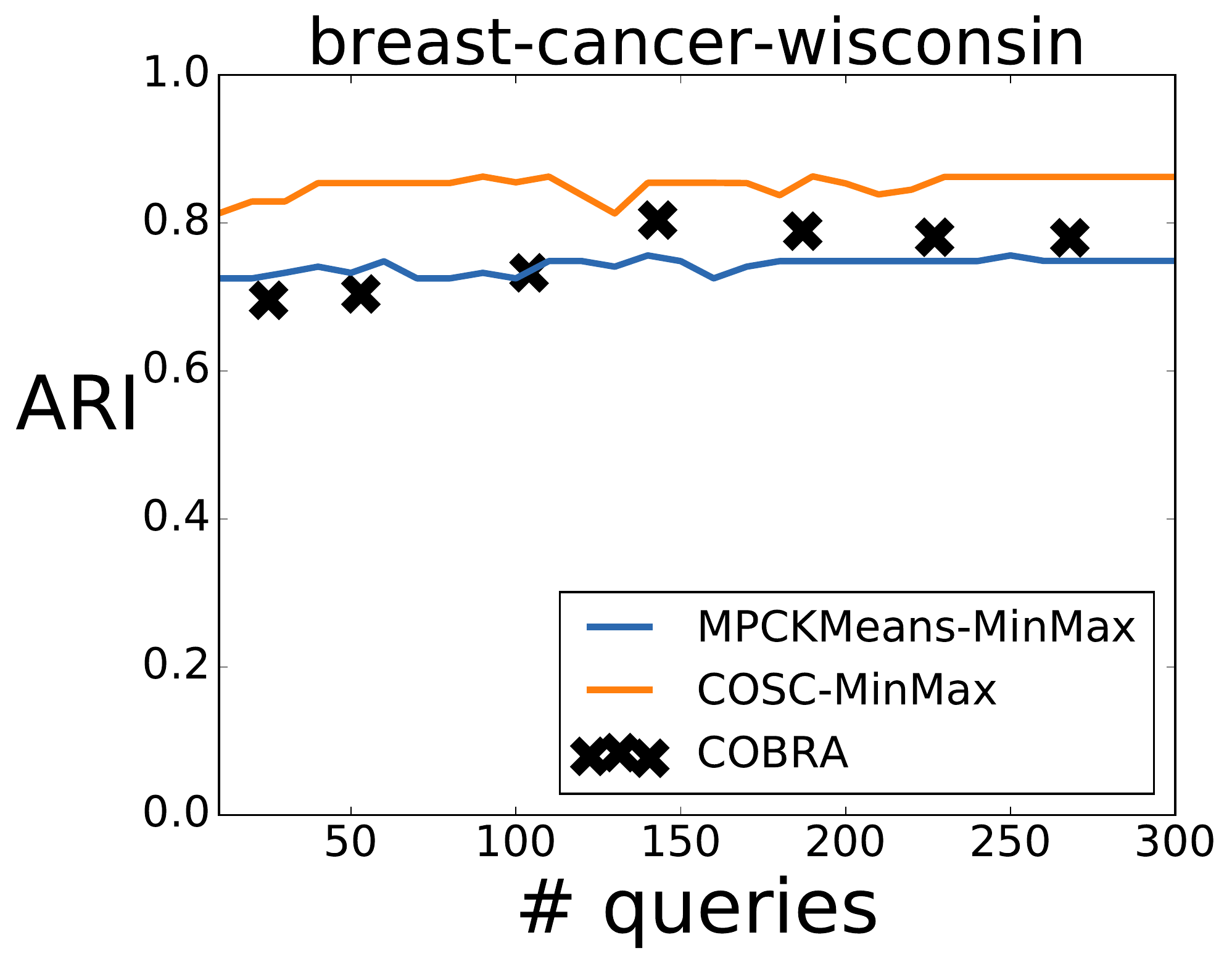}
  \caption{}
  \label{fig:n_constraints_4}
\end{subfigure}\caption{Comparing the clustering qualities for COBRA and its competitors for a wider range of numbers of constraints. For COBRA, each black marker shows the average number of questions that COBRA required for a particular number of super-instances, and its average ARI (over 5-fold CV). We only show the results for the MinMax selection strategy, but the conclusions that are drawn also hold for NPU. The number of super-instances for COBRA is 25, 50, 100 (as in the experiments before), 150, 200, 250, \ldots}\label{fig:more_constraints}
\end{figure*}    

\subsubsection*{Running Competitors with Different Numbers of Queries}
In the previous experiments, the competitors are run with the same number of queries that COBRA required, as for COBRA this cannot be fixed beforehand. One might wonder whether this constitutes an advantage for COBRA, and whether the above conclusions also hold when competitors can be run with different numbers of constraints. To answer this question, we run COBRA with a wider range of super-instances, and its competitors with more numbers of constraints. Figure \ref{fig:more_constraints} shows the results for 4 datasets, but the conclusions that are drawn here also hold for the others. A first conclusion is that for the datasets for which COBRA outperforms its competitors in the experiments discussed above, it also does so for larger numbers of constraints (e.g.\ in Figures \ref{fig:n_constraints_1} and \ref{fig:n_constraints_2}). As such, the results discussed in the previous section are representative. Secondly, clustering quality quickly plateaus for many datasets (e.g.\ in Figure \ref{fig:n_constraints_4}) . This is especially true for MPCKMeans, which can be explained by its strong spherical bias. In contrast, for several datasets both COBRA and COSC produce increasingly better clusterings as more constraints are given (e.g.\ in Figures \ref{fig:n_constraints_1} and \ref{fig:n_constraints_3}). 
   
\subsubsection*{Selecting the Right Number of Clusters}
COBRA does not require specifying the number of clusters $K$ beforehand. Most often, it produces the correct $K$ (i.e.\ the one indicated by the class labels). When it does not, the $K$ it finds is very close to the correct one. In contrast, COSC and MPCKMeans do require specifying $K$. In the experiments discussed above, both of them were given the correct $K$. We have found experimentally that, in the majority of cases, running them with a different $K$ reduces clustering quality, and often by a significant amount. Occasionally a different $K$ improves results, but when this was the case it was typically only by a small margin. These results are omitted from the paper due to lack of space.  

\subsubsection*{Runtime}
\begin{figure}[ht]
\centering
  \centering
  \includegraphics[width=0.7\linewidth]{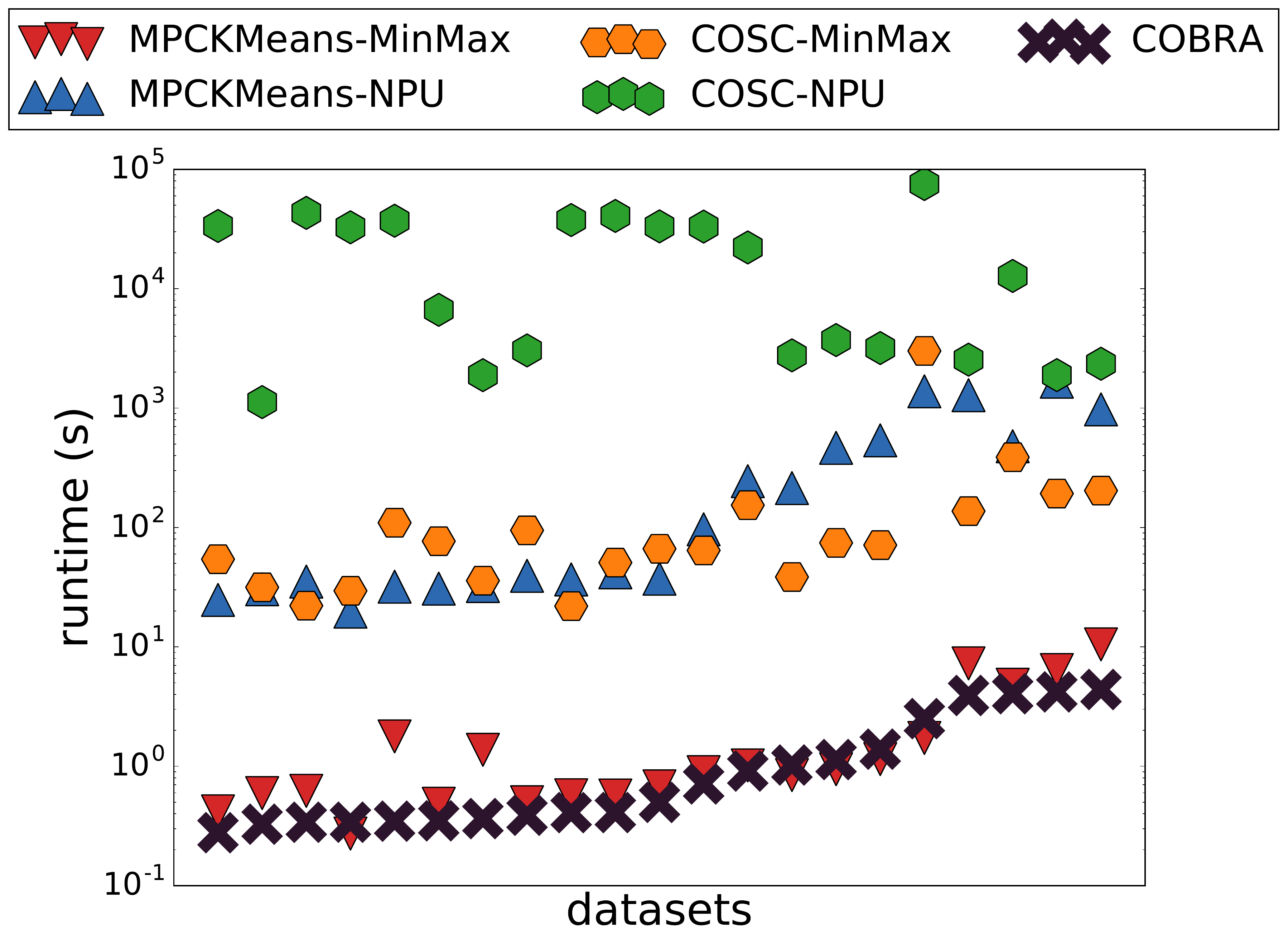}
  \caption{Each dot shows the runtime (averaged over 5 CV folds) of a method on one of the datasets. This figure shows the results for COBRA with 50 super-instances, the figures for 25 and 100 super-instances are comparable. COBRA and MPCKMeans-MinMax are highly scalable and always finish in under 10 seconds. }
  \label{fig:runtimes}
\end{figure}

Figure \ref{fig:runtimes} compares the runtimes of COBRA to those of the competitors. COBRA consists of two steps: constructing super-instances, and grouping them together to form clusters. Both are fast, as K-means is used for the first step and the second step is only applied to the small set of super-instances. As can be seen in Figure \ref{fig:runtimes}, the runtimes of MPCKMeans-MinMax are comparable to those of COBRA, which is not surprising as it is built on K-means. In contrast, COSC-MinMax is significantly more expensive, as it is built on spectral clustering. When used with the NPU selection strategy, both MPCKMeans and COSC become much slower, as NPU requires several runs of the clustering algorithm.

\section{Conclusion}
\label{sec:conclusion}
We have introduced COBRA, an active constraint-based clustering method. Unlike other methods, it is not built as an extension of an existing unsupervised algorithm. Instead, COBRA is inherently constraint-based. With its selection strategy, it aims to maximally exploit constraint transitivity and entailment. Our experiments show that COBRA outperforms the state of the art in terms of clustering quality and runtime, even when the other methods have the advantage of being given the right number of clusters. In future work, we will investigate whether an appropriate number of super-instances can be determined automatically, e.g.\ by incrementally refining them when necessary.  

\section*{Acknowledgements}
Toon Van Craenendonck is supported by the Agency for Innovation by Science and Technology in Flanders (IWT). Sebastijan Duman\v{c}i\'{c} is supported by Research Fund KU Leuven (GOA/13/010).


\bibliographystyle{named}
\bibliography{references}

\end{document}